\documentclass[letterpaper, 10 pt, journal, twoside]{IEEEtran}

\usepackage{times}
\usepackage[pdftex]{graphicx}
\usepackage{amsmath,amssymb,amsopn,amstext,amsfonts}
\usepackage{cancel}
\usepackage[space]{cite}
\usepackage{pdfsync}
\usepackage{balance}
\usepackage{color}
\usepackage{mathtools}
\usepackage{algpseudocode}
\usepackage{algorithm} 
\usepackage{bm}

\usepackage{diagbox}
\usepackage{float}
\usepackage{epstopdf}
\usepackage{pifont}
\usepackage{amsmath}
\usepackage{multirow}
\usepackage{url}
\usepackage{verbatim}
\usepackage[linkcolor=black,citecolor=black,urlcolor=black,colorlinks=true]{hyperref}
\usepackage{booktabs}
\usepackage{graphicx}
\usepackage{subcaption}
\usepackage{threeparttable}  

\graphicspath{{./figures/}}
\DeclareGraphicsExtensions{.png,.jpg,.eps,.pdf}
\IEEEoverridecommandlockouts

\author{Xin Zhou, Zhepei Wang, Hongkai Ye, Chao Xu and Fei Gao
	\thanks{Manuscript received: August 19, 2020; Accepted: October 21, 2020. This paper was recommended for publication by Editor Nancy Amato upon evaluation of the Associate Editor and Reviewers' comments. This work was supported by the Fundamental Research Funds for the Central Universities, under Grant 2020QNA5013. (\textit{Corresponding author: Fei Gao.})}
	\thanks{All authors are with the State Key Laboratory of Industrial Control Technology, and the Institute of Cyber-Systems and Control, Zhejiang University, Hangzhou, 310027, China. {\tt\small \{iszhouxin, wangzhepei, hkye, cxu, and fgaoaa\}@zju.edu.cn}}
	\thanks{Digital Object Identifier (DOI): see top of this page.}
}

\title{EGO-Planner: An ESDF-free Gradient-based Local Planner for Quadrotors}

\begin{document}
%
    
    \maketitle
    
    \begin{abstract}
    	Gradient-based planners are widely used for quadrotor local planning, in which a Euclidean Signed Distance Field (ESDF) is crucial for evaluating gradient magnitude and direction.
        Nevertheless, computing such a field has much redundancy since the trajectory optimization procedure only covers a very limited subspace of the ESDF updating range.
        In this paper, an ESDF-free gradient-based planning framework is proposed, which significantly reduces computation time.
        The main improvement is that the collision term in penalty function is formulated by comparing the colliding trajectory with a collision-free guiding path.
        The resulting obstacle information will be stored only if the trajectory hits new obstacles, making the planner only extract necessary obstacle information.
        Then, we lengthen the time allocation if dynamical feasibility is violated.
        An anisotropic curve fitting algorithm is introduced to adjust higher order derivatives of the trajectory while maintaining the original shape.
        Benchmark comparisons and real-world experiments verify its robustness and high-performance.
        The source code is released as ros packages.
    \end{abstract}

    
    
    \section{Introduction}
    \label{sec:introduction}
    \IEEEPARstart{I}{n} recent years, the emergence of quadrotor online planning methods has greatly pushed the boundary of aerial autonomy, making drones fly out of laboratories and appear in numerous real-world applications.
    Among these methods, gradient-based ones, which smooth a trajectory and utilize the gradient information to improve its clearance, have shown great potential and gain more and more popularity\cite{quan2020survey}.  
    
    Traditionally, gradient-based planners rely on a pre-built ESDF map to evaluate the gradient magnitude and direction, and use numerical optimization to generate a local optimal solution.
    Although the optimization programs enjoy fast convergence, they suffer a lot from constructing the required ESDF beforehand.
    As the statistics (\textit{TABLE II} from EWOK\cite{Usenko2017ewok})  states, the ESDF computation takes up to about 70\% of total processing time for conducting local planning. 
    Therefore, we can safely claim that, building ESDF has become the bottleneck of gradient-based planners, preventing the method from being applied to resource-limited platforms.
    
\begin{figure}[t]
        \centering
        \includegraphics[width=1\linewidth]{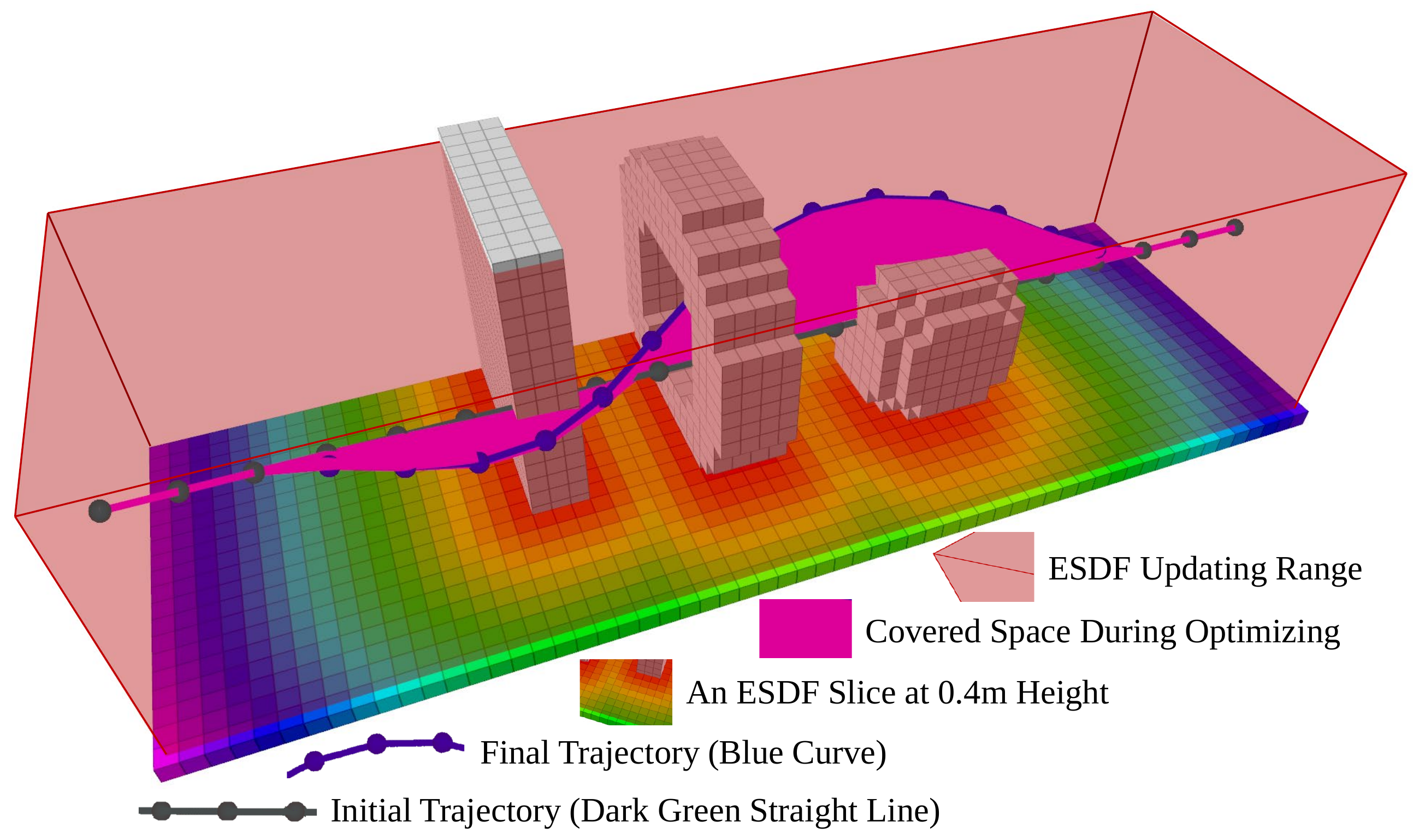}
        \captionsetup{font={small}}
        \caption{ Trajectory during optimizing just covers a very limited space of the ESDF updating range. }
        \label{pic:esdf}
        \vspace{-0.8cm}
    \end{figure}

    Though ESDF is widely used, few works analyze its necessity. 
    Typically, there are two ways to build an ESDF.
    As detailed in Sec.\ref{sec:related_work}, methods can be categorized as the incremental global updating~\cite{han2019fiesta} ones, and the batch local calculation~\cite{felzenszwalb2012distance} ones. 
    However, neither of them focuses on the trajectory itself. 
    Consequently, too much computation is spent on calculating ESDF values that make no contribution to the planning.
    In other words, current ESDF-based methods do not serve the trajectory optimization solely and directly.
    As shown in Fig.\ref{pic:esdf}, for a general autonomous navigation scenario where the drone is expected to avoid collisions locally, the trajectory covers only a limited space of the ESDF updating range.
    In practice, although some handcrafted rules can decide a slim ESDF range, they lack theoretical rationality and still induce unnecessary computations.
    
    In this paper, we design an \textbf{E}SDF-free \textbf{G}radient-based l\textbf{O}cal planning framework called \textbf{EGO}, and we incorporate careful engineering considerations to make it lightweight and robust.
    The proposed algorithm is composed of a gradient-based spline optimizer and a post-refinement procedure. 
    Firstly, we optimize the trajectory with smoothness, collision, and dynamical feasibility terms.
    Unlike traditional approaches that query pre-computed ESDF, we model the collision cost by comparing the trajectory inside obstacles with a guiding collision-free path.
    We then project the forces onto the colliding trajectory and generate estimated gradient to wrap the trajectory out of obstacles.
    During the optimization, the trajectory will rebound a few times between nearby obstacles and finally terminate in a safe region. 
    In this way, we only calculate the gradient when necessary, and avoid computing ESDF in regions irrelevant to the local trajectory.
    If the resulted trajectory violates dynamical limits, which is usually caused by unreasonable time allocation, the refinement process is activated. 
    During the refinement, trajectory time is reallocated when the limits are exceeded.
    With the enlarged time allocation, a new B-spline that fits the previous dynamical infeasible one while balancing the feasibility and fitting accuracy is generated.
    To improve robustness, the fitting accuracy is modeled anisotropically with different penalties on axial and radial directions.
    
    To the best knowledge of us, this method is the first to achieve gradient-based local planning without an ESDF. 
    Compared to existing state-of-the-art works, the proposed method generates safe trajectories with comparable smoothness and aggressiveness, but lower computation time of over an order of magnitude by omitting the ESDF maintenance.
    We perform comprehensive tests in simulation and real-world to validate our method. 
    Contributions of this letter are:
    \begin{enumerate}
        \item 
        We propose a novel and robust gradient-based quadrotor local planning method, which evaluates and projects gradient information directly from obstacles instead of a pre-built ESDF.
        \item 
        We propose a lightweight yet effective trajectory refinement algorithm, which generates smoother trajectories by formulating the trajectory fitting problem with anisotropic error penalization.
        \item 
        We integrate the proposed method into a fully autonomous quadrotor system, and release our software for the reference of the community\footnote{https://github.com/ZJU-FAST-Lab/ego-planner}.  
    \end{enumerate}
    
    \begin{figure}[t]
        \centering
        \includegraphics[width=1\linewidth]{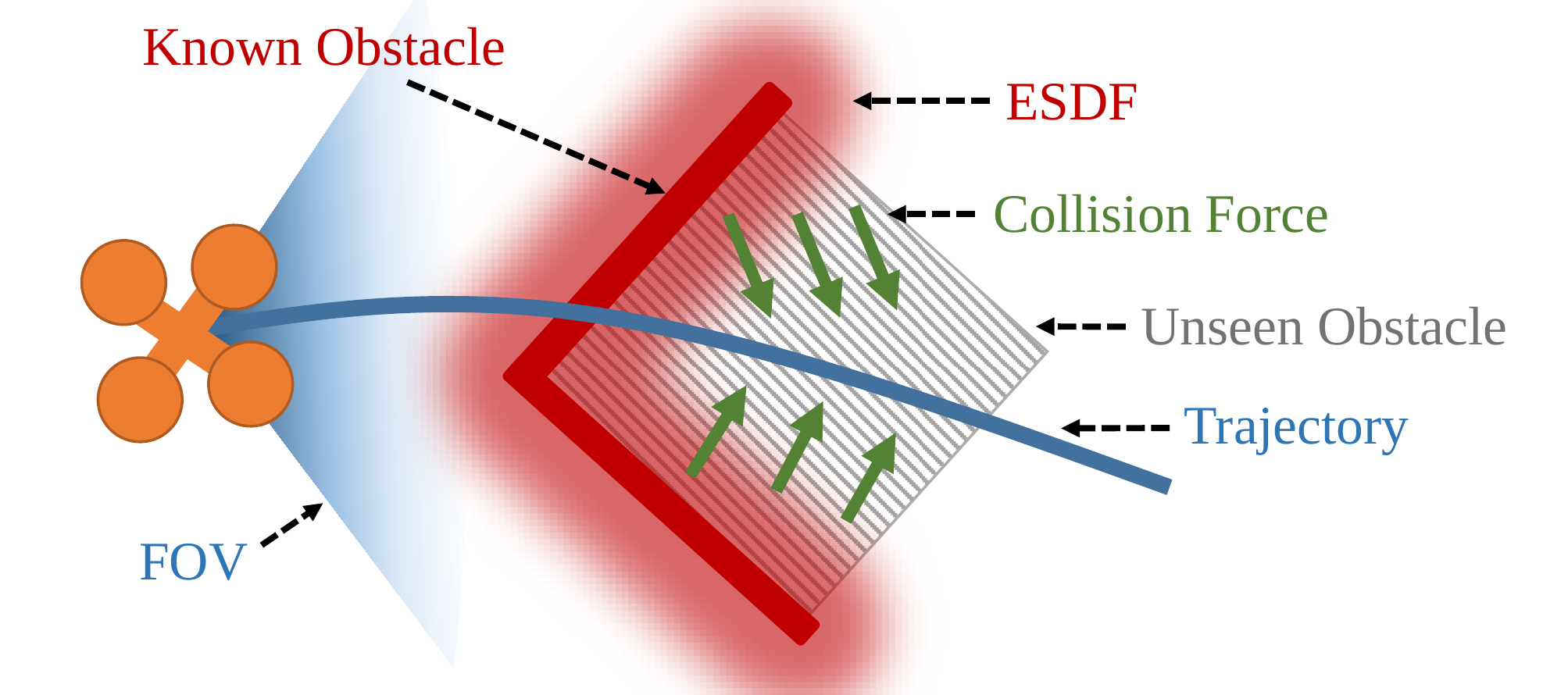}
        \captionsetup{font={small}}
        \caption{ The trajectory gets stuck into a local minimum, which is very common since the camera has no vision of the back of the obstacle.}
        \label{pic:local_min}
        \vspace{-1.0cm}
    \end{figure}
    
    \section{Related Work}
    \label{sec:related_work}
    \subsection{Gradient-based Motion Planning}
    Gradient-based motion planning is the mainstream for UAV local trajectory generation, which formulates the problem as unconstrained nonlinear optimization. 
    ESDF is first introduced in robotic motion planning by Ratliff et al.\cite{ratliff2009chomp}. Utilizing its abundant gradient information, many planning frameworks directly optimize trajectories in the configuration space.
    Nevertheless, optimizing the trajectory in discrete-time\cite{ratliff2009chomp, kalakrishnan2011stomp} is not suitable for drones, because it is much more sensitive to dynamical constraints.
    Thereby, \cite{oleynikova2016continuous} proposes a continuous-time polynomial trajectory optimization method for UAV planning. 
    However, the involved integral of the potential function causes a heavy computation burden.
    Besides, the success rate of this method is around $70\%$, even with random restarts.
    For these drawbacks, \cite{Usenko2017ewok} introduces a B-spline parameterization of the trajectory which takes good advantage of the convex hull property.
    In \cite{gao2017gradient}, the success rate is significantly increased by finding a collision-free initial path as the front-end.
    Moreover, the performance is further improved when the generation of the initial collision-free path takes into account kinodynamic constraints \cite{ding2019an, zhou2019robust}.
    Zhou et al.\cite{zhou2020raptor} incorporate perception awareness to make the system more robust.
    Among the above approaches, ESDF plays a vital role in evaluating distance with gradient magnitude and direction to nearby obstacles. 
    
    \subsection{Euclidean Signed Distance Field (ESDF)}
    ESDF has long been used to construct objects from noisy sensor data for over two decades \cite{curless1996volumetric}, and revive interests in robotics motion planning since \cite{ratliff2009chomp}.
    Felzenszwalb et al. \cite{felzenszwalb2012distance} propose an envelope algorithm that reduces the time complexity of ESDF construction to $O(n)$ with $n$ denoted as voxel numbers.
    This algorithm is not suitable for incremental building of ESDF, while dynamic updating of the field is often needed during quadrotor flight.
    To solve this problem, Oleynikova \cite{oleynikova2017voxblox} and Han \cite{han2019fiesta} propose incremental ESDF generation methods, namely \textit{Voxblox} and \textit{FIESTA}.
    Although these methods are highly efficient in dynamic updating cases, the generated ESDF almost always contains redundant information that may not be used in the planning procedure at all. 
    As is shown in Fig.\ref{pic:esdf}, this trajectory only sweeps over a very limited subspace of the whole ESDF updating range. 
    Therefore, it is valuable to design a more intelligent and lightweight method, instead of maintaining the whole field.
    
    \begin{figure*}[t]
        \centering
        \begin{subfigure}{0.333\linewidth}
            \includegraphics[width=1\linewidth]{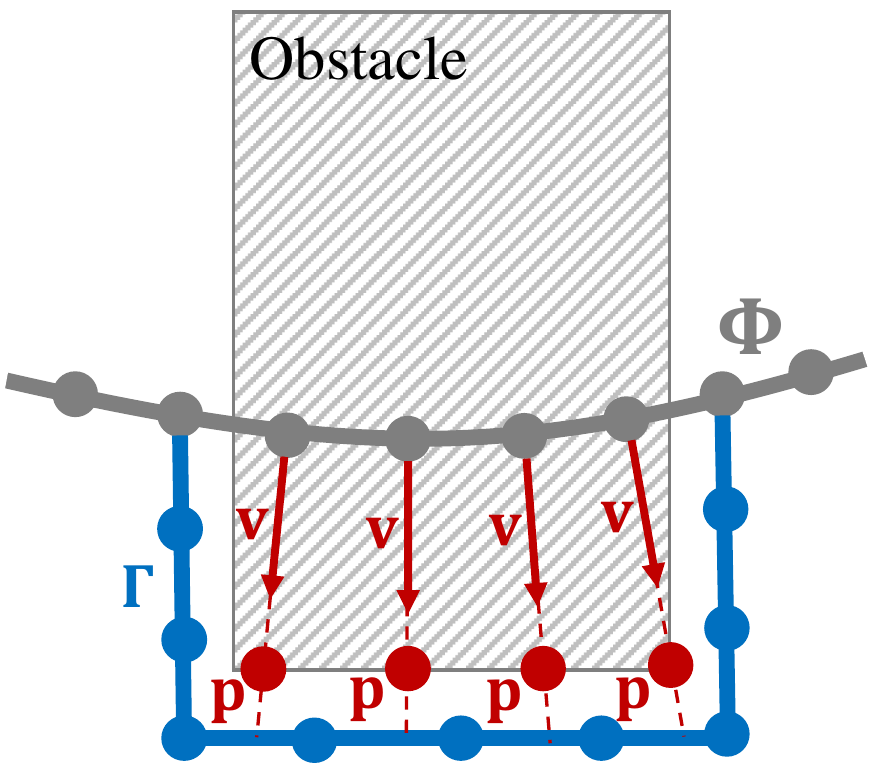}
            \captionsetup{font={small}}
            \caption{$\{\mathbf{p},\mathbf{v}\}$ Pairs}
            \label{pic:2d_p_v_pairs}
        \end{subfigure}
        \begin{subfigure}{0.22\linewidth}
            \includegraphics[width=1\linewidth]{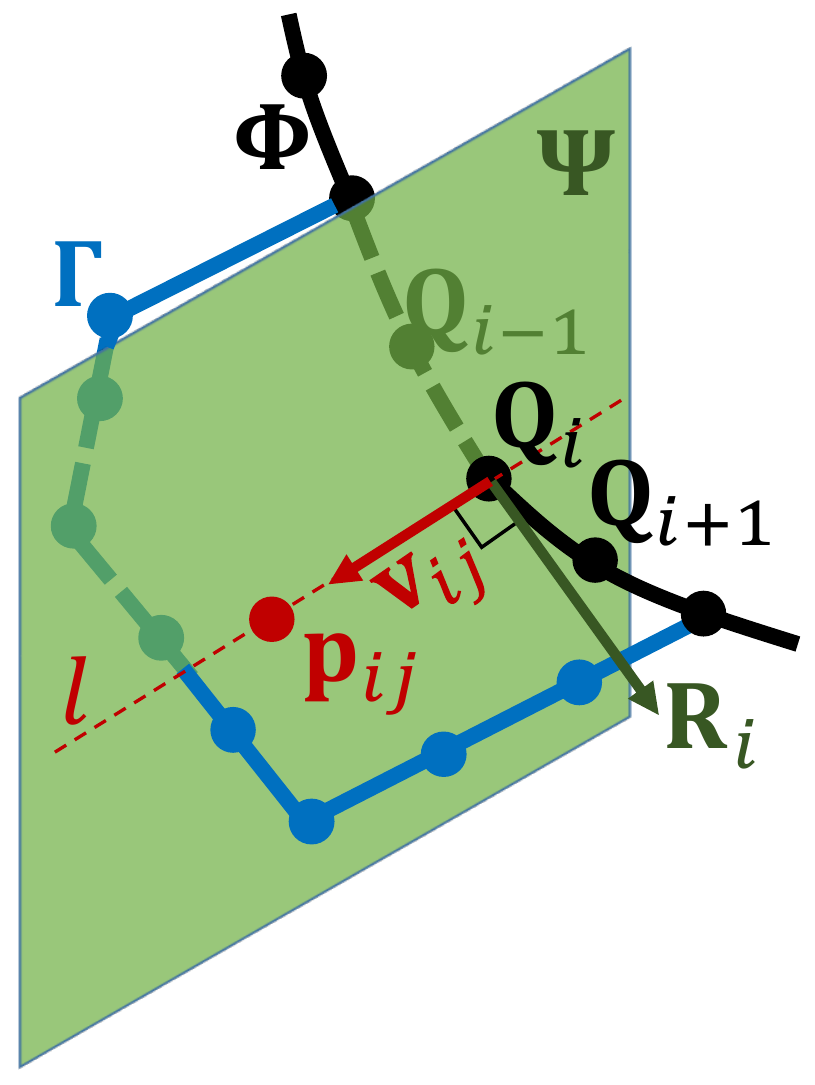}
            \captionsetup{font={small}}
            \caption{$\mathbf{p},\mathbf{v}$ Generation}
            \label{pic:3d_p_v_pairs}
        \end{subfigure}
        \begin{subfigure}{0.34\linewidth}
            \includegraphics[width=1\linewidth]{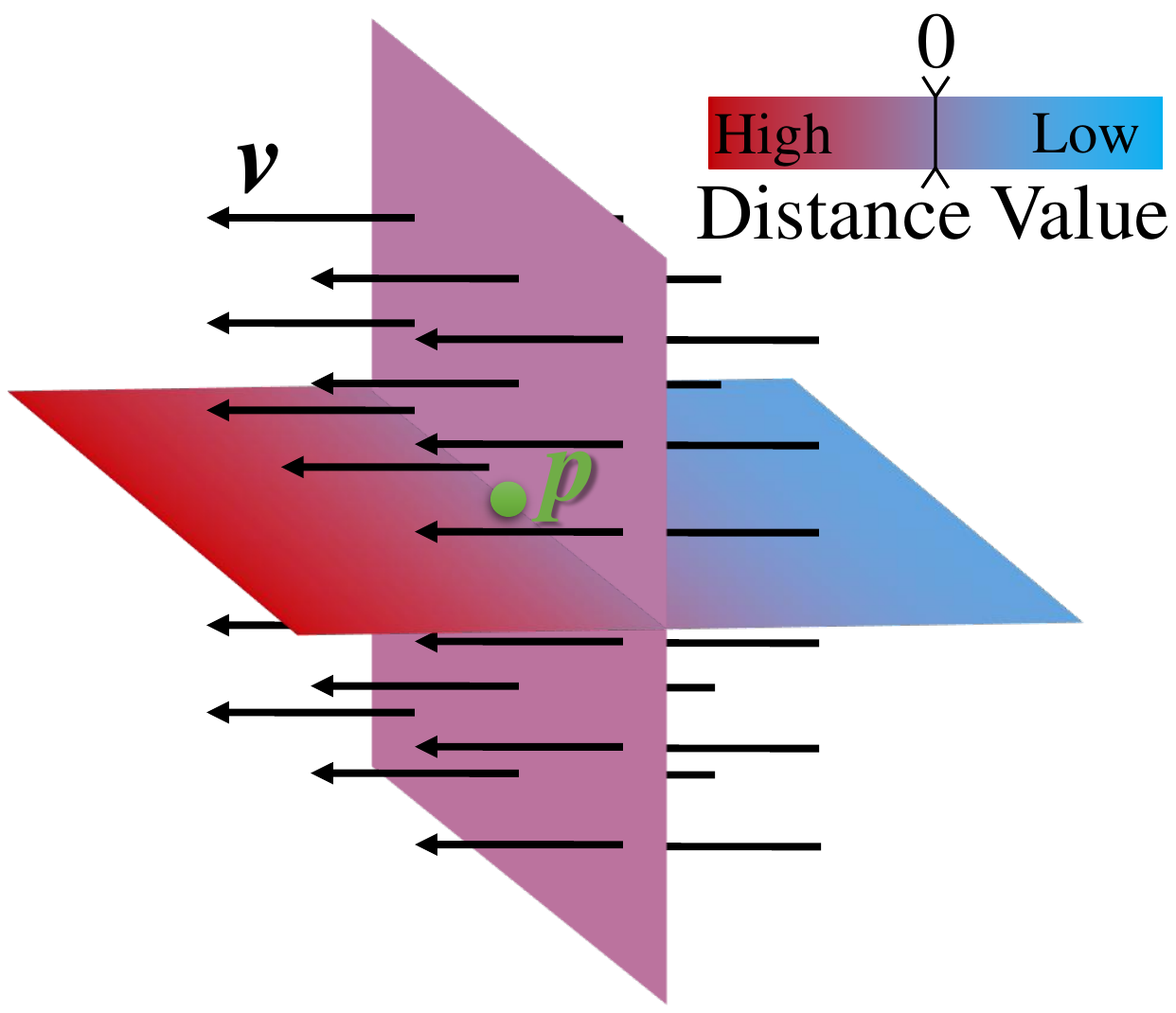}
            \captionsetup{font={small}}
            \caption{Distance Field of A $\{\mathbf{p},\mathbf{v}\}$ Pair}
            \label{pic:3d_dis_field}	
        \end{subfigure}
    	\captionsetup{font={small}}
        \caption{a) A trajectory $\mathbf{\Phi}$ passing through an obstacle generates several $\{\mathbf{p},\mathbf{v}\}$ pairs for control points. $\mathbf{p}$ are the points at the obstacle surface and $\mathbf{v}$ are unit vectors pointing from control points to $\mathbf{p}$. b) A plane $\mathbf{\Psi}$ which is perpendicular to a tangent vector $\mathbf{R}_i$ intersects $\mathbf{\Gamma}$ forming a line $l$, from which a $\{\mathbf{p},\mathbf{v}\}$ pair is determined. c) Slice visualization of distance field definition $d_{ij}=(\mathbf{Q}_{i}-\mathbf{p}_{ij}) \cdot \mathbf{v}_{ij}$. The color indicates the distance and the arrows are identical gradients equal to $\mathbf{v}$. $\mathbf{p}$ is at the zero distance plane. }
        \label{pic:p_v_visualization}
        \vspace{-0.5cm}
    \end{figure*}

	\setlength{\textfloatsep}{0pt}
	\begin{algorithm}[t]
		\caption{CheckAndAddObstacleInfo}
		\label{alg:StoreObstacleInfo}
		\begin{algorithmic}[1]
			\State \textbf{Notation}: Environment $\mathcal{E}$, Control Points Struct $\mathbf{Q}$, Anchor Points $\mathbf{p}$, Repulsive Direction Vector $\mathbf{v}$, Colliding Segments $\mathbf{S}$
			\State Input: $\mathcal{E}$, $\mathbf{Q}$
			\For{$\mathbf{Q}_i$ in $\mathbf{Q}$}
			\If{FindConsecutiveCollidingSegment($\mathbf{Q}_i$)}
			\State $\mathbf{S}$.push\_back(GetCollisionSegment())
			\EndIf
			\EndFor
			\For{$\mathbf{S}_i$ in $\mathbf{S}$}
			\State $\mathbf{\Gamma} \leftarrow$ PathSearch($\mathcal{E}$, $\mathbf{S_i}$)
			\For{$\mathbf{S}_i.\text{begin} \leq j \leq \mathbf{S}_i.\text{end}$}
			\State $\{\mathbf{p},\mathbf{v}\} \leftarrow$ Find\_p\_v\_Pairs($\mathbf{Q}_j, \mathbf{\Gamma}$)
			\State $\mathbf{Q}_j$.push\_back($\{\mathbf{p},\mathbf{v}\}$) 
			\EndFor
			\EndFor
		\end{algorithmic}
	\end{algorithm}
    
    \section{Collision Avoidance Force Estimation}
    \label{sec:distance_estimate}
    
    In this paper, the decision variables are control points $\mathbf{Q}$ of a B-spline curve.
    Each $\mathbf{Q}$ possesses its own environment information independently.
    Initially, a naive B-spline curve $\mathbf{\Phi}$ satisfying terminal constraints is given, regardless of collision.
    Then, the optimization procedure starts. 
    For each colliding segment detected in an iteration, a collision-free path $\mathbf{\Gamma}$ is generated. 
    Each control point $\mathbf{Q}_{i}$ of the colliding segment, after that, will be assigned an anchor point $\mathbf{p}_{ij}$ at the obstacle surface with a corresponding repulsive direction vector $\mathbf{v}_{ij}$, as shown in Fig.\ref{pic:2d_p_v_pairs}.
    Denote by $i \in \mathbb{N}_+$ the index of control points, and $j \in \mathbb{N}$ the index of $\{\mathbf{p},\mathbf{v}\}$ pair. 
    Note that each $\{\mathbf{p},\mathbf{v}\}$ pair only belongs to one specific control point. For brevity, we omit the subscript $ij$ without causing ambiguity.
    The detailed $\{\mathbf{p},\mathbf{v}\}$ pair generation procedure in this paper is summarized in Alg.\ref{alg:StoreObstacleInfo} and is illustrated in Fig.\ref{pic:3d_p_v_pairs}.
    Then the obstacle distance from $\mathbf{Q}_{i}$ to the $j^{th}$ obstacle is defined as 
    \begin{equation}
    d_{ij}=(\mathbf{Q}_{i}-\mathbf{p}_{ij}) \cdot \mathbf{v}_{ij}.
    \end{equation}
    
    In order to avoid duplicative $\{\mathbf{p},\mathbf{v}\}$ pair generation before the trajectory escapes from the current obstacle during the first several iterations, we adopt a criterion that considers an obstacle which the control point $\mathbf{Q}_i$ lies in as newly discovered, only if the current $\mathbf{Q}_i$ satisfies $d_{ij}>0$ for all valid $j$.
    Besides, this criterion allows only necessary obstacles that contribute to the final trajectory to be taken into optimization. 
    Thus, the operation time is significantly reduced.
    
    To incorporate necessary environmental awareness into the local planner, we need to explicitly construct an objective function that keeps the trajectory away from obstacles. 
    ESDF provides this vital collision information but with the price of a heavy computation burden.
    In addition, as shown in Fig.\ref{pic:local_min}, ESDF-based planners can easily fall into a local minimum and fail to escape from obstacles, due to the insufficient or even wrong information from ESDF.
    To avoid such situations, an additional front-end is always needed to provide a collision-free initial trajectory.
    The above methodology outperforms ESDF in providing the vital information for collision avoidance, since the explicitly designed repulsive force can be fairly effective regarding various missions and environments.
    Moreover, the proposed method has no requirement for collision-free initialization.
    
    \section{Gradient-Based Trajectory Optimization}
    \label{sec:b_spline_opt}
    
    \subsection{Problem Formulation}
    \label{sec:problem_formulation}
    
    In this paper, the trajectory is parameterized by a uniform B-spline curve $\mathbf{\Phi}$, which is uniquely determined by its degree $p_b$, $N_c$ control points $\left\lbrace \mathbf{Q}_1, \mathbf{Q}_2, \cdots, \mathbf{Q}_{N_c} \right\rbrace $, and a knot vector $\left\{ t_1, t_2, \cdots, t_M \right\}$, where $\mathbf{Q}_i \in \mathbb{R}^3$, $t_m \in \mathbb{R}$ and $M=N_c+p_b$.
    For simplicity and efficiency of trajectory evaluation, the B-spline used in our method is uniform, which means each knot is separated by the same time interval $\triangle t=t_{m+1}-t_m$ from its predecessor.
    The problem formulation in this paper is based on the current state-of-the-art quadrotor local planning framework Fast-Planner \cite{boyu2019ral}.
    
    \begin{figure}[t]
    	\centering
    	\includegraphics[width=0.9\linewidth]{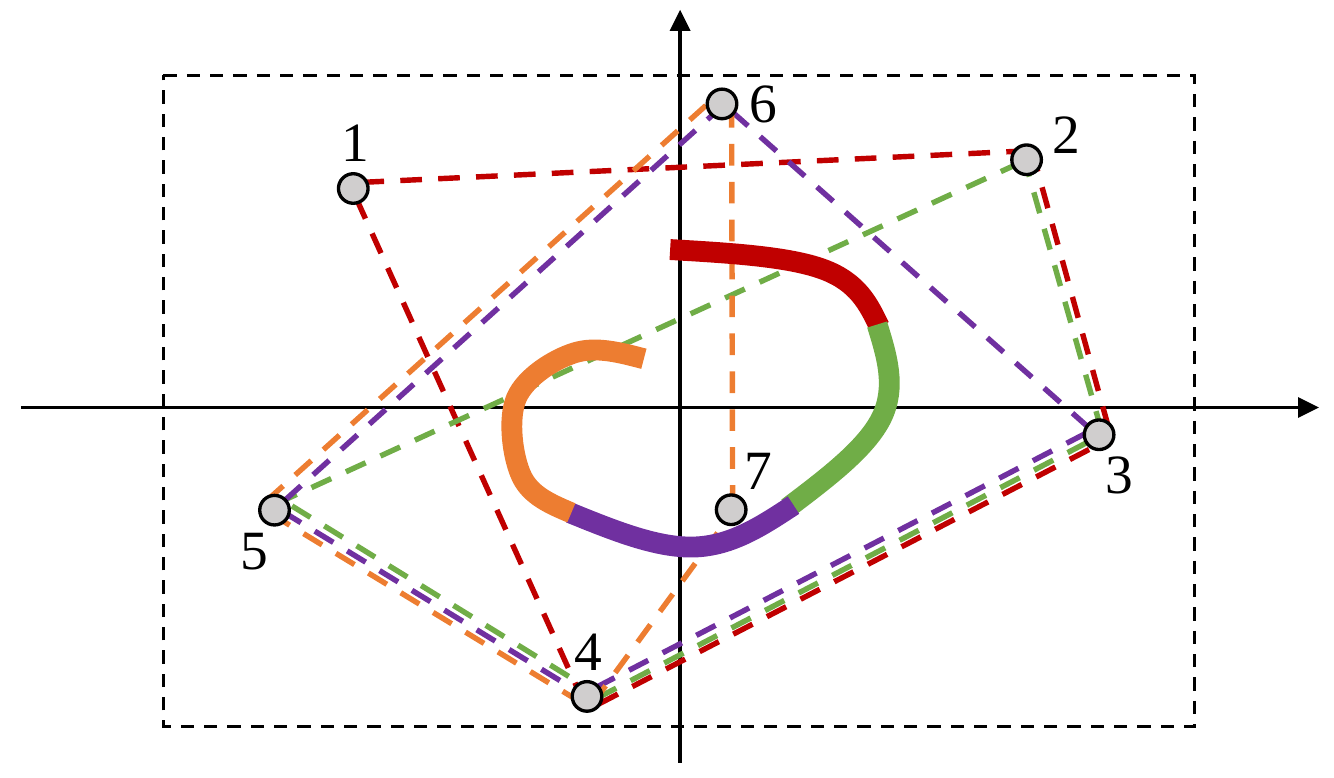}
    	\captionsetup{font={small}}
    	\caption{ Convex hull property of the B-spline curve. Gray points represent control points. The whole curve stays inside the feasibility bounding box(black dotted box) as long as all the control points are in that box. Without loss of generality, each convex hull consists of four vertexes. }
    	\label{pic:cvx_hull_prot}
    \end{figure}
    
    B-spline enjoys convex hull property.
    This property indicates that a single span of a B-spline curve is merely controlled by $p_b+1$ successive control points and lies within the convex hull of these points.
    For example, a span within $(t_i, t_{i+1})$ lies inside the convex hull formed by $\{\mathbf{Q}_{i-p_b}, \mathbf{Q}_{i-p_b+1}, \cdots, \mathbf{Q}_i\}$.
    Another property is that the $k^{th}$ derivative of a B-spline is still a B-spline with order $p_{b,k}=p_b-k$.
    Since $\triangle t$ is identical alone $\mathbf{\Phi}$, the control points of the velocity $\mathbf{V}_i$, acceleration $\mathbf{A}_i$, and jerk $\mathbf{J}_i$ curves are obtained by 
    \begin{equation}
    \label{equ:v_a_j}
    \mathbf{V}_i=\frac{\mathbf{Q}_{i+1}-\mathbf{Q}_i}{\triangle t}, \ \mathbf{A}_i=\frac{\mathbf{V}_{i+1}-\mathbf{V}_i}{\triangle t}, \ \mathbf{J}_i=\frac{\mathbf{A}_{i+1}-\mathbf{A}_i}{\triangle t}.
    \end{equation}
    
    We follow the work of \cite{MelKum1105} to plan the control points $\mathbf{Q} \in \mathbb{R}^3$ in a reduced space of  \textit{differentially flat outputs}.
    The optimization problem is then formulated as follows:
    \begin{equation}
    \label{equ:obj_fun}
    \mathop{min}_{\mathbf{Q}} J = \lambda_s J_s + \lambda_c J_c + \lambda_d J_d,
    \end{equation}
    where $J_s$ is the smoothness penalty, $J_c$ is for collision, and $J_d$ indicates feasibility.
    $\lambda_s, \lambda_c, \lambda_d$ are weights for each penalty terms.
    
    \subsubsection{Smoothness penalty}
    \label{sec::smoothness_penalty}
    
    In \cite{Usenko2017ewok}, the smoothness penalty is formulized as the time integral over square derivatives of the trajectory (acceleration, jerk, etc.). 
    In \cite{zhou2019robust}, only geometric information of the trajectory is taken regardless of time allocation.
    In this paper, we combine both methods to penalize squared acceleration and jerk without time integration.
    
    Benefiting from the convex hull property, minimizing the control points of second and third order derivatives of the B-spline trajectory is sufficient to reduce these derivatives along the whole curve.
    Therefore, the smoothness penalty function is formulated as
    \begin{equation}
    J_s = \sum_{i=1}^{N_c-1}\|\mathbf{A}_i\|_2^2 + \sum_{i=1}^{N_c-2}\|\mathbf{J}_i\|_2^2,
    \end{equation}
    which minimizes high order derivatives, making the whole trajectory smooth. 
    
    \subsubsection{Collision penalty}
    \label{sec::colli_fun}
    
    Collision penalty pushes control points away from obstacles.
    This is achieved by adopting a safety clearance $s_f$ and punishing control points with $d_{ij}<s_f$.
    In order to further facilitate optimization, we construct a twice continuously differentiable penalty function $j_c$ and suppress its slope as $d_{ij}$ decreases, which yields the piecewise function 
    \begin{gather}
    j_c(i,j) = \left\{
    \begin{aligned}
    & 0 								& (c_{ij} \leq 0) \\
    & c_{ij}^3 							& (0 < c_{ij} \leq s_f)\\
    & 3s_f c_{ij}^2-3s_f^2 c_{ij}+s_f^3 & (c_{ij} > s_f)
    \end{aligned}
    \right.,\\
    c_{ij}=s_f-d_{ij}, \qquad \qquad \qquad \qquad \qquad \qquad \quad \notag
    \end{gather}
    where $j_c(i,j)$ is the cost value produced by $\{\mathbf{p},\mathbf{v}\}_j$ pairs on $\mathbf{Q}_i$.
    The cost on each $\mathbf{Q}_i$ is evaluated independently and accumulated from all corresponding $\{\mathbf{p},\mathbf{v}\}_j$ pairs. 
    Thus, a control point obtains a higher trajectory deformation weight if it discovers more obstacles.
    Specifically, the cost value added to the $i^{th}$ control point is $j_c(\mathbf{Q}_i)=\sum_{j=1}^{N_p} j_c(i,j)$, $N_p$ is the number of $\{\mathbf{p},\mathbf{v}\}_j$ pairs belonging to $\mathbf{Q}_i$.
    Combining costs on all $\mathbf{Q}_i$ yields the total cost $J_c$, i.e.,
    \begin{equation}
    J_c = \sum_{i=1}^{N_c} j_c(\mathbf{Q}_i).
    \end{equation}
    
    Unlike traditional ESDF-based methods \cite{Usenko2017ewok, zhou2019robust}, which compute gradient by trilinear interpolation on the field, we obtain gradient by directly computing the derivative of $J_c$ with respect to $\mathbf{Q}_i$, which gives
    \begin{equation}
    \frac{\partial J_c}{\partial \mathbf{Q}_i} = \sum_{i=1}^{N_c}\sum_{j=1}^{N_p}\mathbf{v}_{ij}\left\{
    \begin{aligned}
    & 0 					& (c_{ij} \leq 0) \\
    & -3 c_{ij}^2  			& (0 < c_{ij} \leq s_f)\\
    & -6s_f c_{ij}+3s_f^2 & (c_{ij} > s_f)
    \end{aligned}
    \right..
    \end{equation}
    
    \subsubsection{Feasibility penalty}
    \label{sec::feasibility_penalty}
    
    Feasibility is ensured by restricting the higher order derivatives of the trajectory on every single dimension, i.e., applying $|\mathbf{\Phi}^{(k)}_r(t)| < \mathbf{\Phi}^{(k)}_{r,max}$ for all $t$, where $r \in \{x, y, z\}$ indicates each dimension.
    Thanks to the convex hull property, constraining derivatives of the control points is sufficient for constraining the whole B-spline.
    Therefore, the penalty function is formulated as
    \begin{equation}
    J_d=\sum_{i=1}^{N_c} w_v F(\mathbf{V}_i) + \sum_{i=1}^{N_c-1} w_a F(\mathbf{A}_i) + \sum_{i=1}^{N_c-2} w_j F(\mathbf{J}_i),
    \end{equation}
    where $w_v, w_a, w_j$ are weights for each terms and $F(\cdot)$ is a twice continuously differentiable metric function of higher order derivatives of control points.
    \begin{equation}
    F(\mathbf{C}) = \mathop{\sum}_{r=x,y,z} f(c_r),
    \end{equation}
    \begin{equation}
    f(c_r) = \left\{
    \begin{aligned}
    & a_1 c_r^2 + b_1 c_r + c_1  & (c_r \leq -c_{j} ) \\
    & (-\lambda c_{m}-c_r)^3   & (-c_{j} < c_r < -\lambda c_{m} ) \\
    & 0						 & ( -\lambda c_{m} \leq c_r \leq  \lambda c_{m}) \\
    & (c_r - \lambda c_{m})^3  & (\lambda c_{m} < c_r < c_{j} ) \\
    & a_2 c_r^2 + b_2 c_r + c_2  & (c_r \geq c_{j} )
    \end{aligned}
    \right.,
    \end{equation}
    where $c_r \in \mathbf{C} \in \{\mathbf{V}_{i}, \mathbf{A}_{i}, \mathbf{J}_{i}\}$, $a_1,b_1,c_1,a_2,b_2,c_2$ are chosen to meet the second-order continuity, $c_m$ is the derivative limit, $c_j$ is the splitting points of the quadratic interval and the cubic interval. 
    $\lambda<1-\epsilon$ is an elastic coefficient with $\epsilon\ll1$ to make the final results meet the constraints, since the cost function is a tradeoff of all weighted terms.

\subsection{Numerical Optimization}
\label{sec::problem_optimization}

The formulated problem in this paper features in two aspects.
Firstly, the objective function $J$ alters adaptively according to the newly found obstacles. 
It requires the solver to be able to restart fast.
Secondly, quadratic terms dominate the formulation of the objective function, making $J$ approximate quadratic.
It means that the utilization of Hessian information can significantly accelerate the convergence.
However, obtaining the exact inverse Hessian is prohibitive in real-time applications since it consumes nonnegligible massive computation. 
To circumvent this, quasi-Newton methods that approximate the inverse Hessian from gradient information are adopted.

Since the performance of a solver is problem dependent, we compare three algorithms belonging to quasi-Newton methods.
They are Barzilai-Borwein method\cite{barzilai1988two} which is capable of fast restart with most crude Hessian estimation, truncated Newton method \cite{steiha1983truncatednewton} which estimates Hessian by adding multiple tiny perturbations to a given state, L-BFGS method \cite{liu1989limited} which approximates Hessian from previous objective function evaluations but requires a serial of iterations to reach a relatively accurate estimation.
Comparison in Sec.\ref{sec:solver_comp} states that L-BFGS outperforms the other two algorithms with appropriately selected memory size, balancing the loss of restart and the accuracy of inverse Hessian estimation. 
This algorithm is briefly explained as follows.
For an unconstrained optimization problem $\mathop{min}_{\mathbf{x} \in {\mathbb{R}^n}} f(\mathbf{x})$, the updating for $\mathbf{x}$ follows the approximated Newton step
\begin{equation}
\label{equ:X}
\mathbf{x}_{k+1}=\mathbf{x}_k- \alpha_k \mathbf{H}_k \nabla \mathbf{f}_k,
\end{equation}
where $\alpha_k$ is the step length and $\mathbf{H}_k$ is updated at every iteration by means of the formula
\begin{equation}
\label{equ:H_k}
\mathbf{H}_{k+1} = \mathbf{V}_k^T \mathbf{H}_{k} \mathbf{V}_k + \rho_k \mathbf{s}_k \mathbf{s}_k^T,
\end{equation}
where $\rho_k = (\mathbf{y}_k^T \mathbf{s}_k)^{-1}, \mathbf{V}_k = \mathbf{I} - \rho_k \mathbf{y}_k \mathbf{s}_k^T, \mathbf{s}_k = \mathbf{x}_{k+1} - \mathbf{x}_k $ and $ \mathbf{y}_k = \nabla \mathbf{f}_{k+1} - \nabla \mathbf{f}_{k} $.

Here $\mathbf{H}_k$ is not calculated explicitly. 
The algorithm right multiplies $\nabla \mathbf{f}_k$ to Equ.\ref{equ:H_k} and recursively expands for $m$ steps and then yields the efficient two-loop recursion updating method\cite{barzilai1988two}, resulting in linear time/space complexity.
The weight of Barzilai-Borwein step is used as the initial inverse Hessian $\mathbf{H}_{k}^0$ for L-BFGS updating, which is 
\begin{equation}
\mathbf{H}_{k}^0 = \frac{\mathbf{s}_{k-1}^T \mathbf{y}_{k-1}}{ \mathbf{y}_{k-1}^T \mathbf{y}_{k-1} } \mathbf{I} \ or \ \frac{\mathbf{s}_{k-1}^T \mathbf{s}_{k-1}}{ \mathbf{s}_{k-1}^T \mathbf{y}_{k-1} } \mathbf{I}.
\end{equation}
A monotone line search under strong Wolfe condition is used to enforce convergence.
	
\section{Time Re-allocation and Trajectory Refinement}
\label{sec:time-allocation}
Allocating an accurate time profile before the optimization is unreasonable, since the planner knows no information about the final trajectory then. 
Therefore, an additional time re-allocation procedure is vital to ensure dynamical feasibility.
Previous works \cite{gao2020teach, zhou2019robust} parameterize the trajectory as a non-uniform B-spline and iteratively lengthen a subset of knot spans when some segments exceed derivative limits.

However, one knot span $\triangle t_n$ influences multiple control points and vice versa, leading to high-order discontinuity to the previous trajectory when adjusting knot spans near the start state.
In this section, a uniform B-spline trajectory $\mathbf{\Phi}_{f}$ is re-generated with reasonable time re-allocation according to the safe trajectory $\mathbf{\Phi}_{s}$ from \ref{sec:b_spline_opt}.
Then, an anisotropic curve fitting method is proposed to make $\mathbf{\Phi}_{f}$ freely optimize its control points to meet higher order derivative constraints while maintaining a nearly identical shape to $\mathbf{\Phi}_{s}$.


\begin{figure}[t]
	\centering
	\includegraphics[width=1\linewidth]{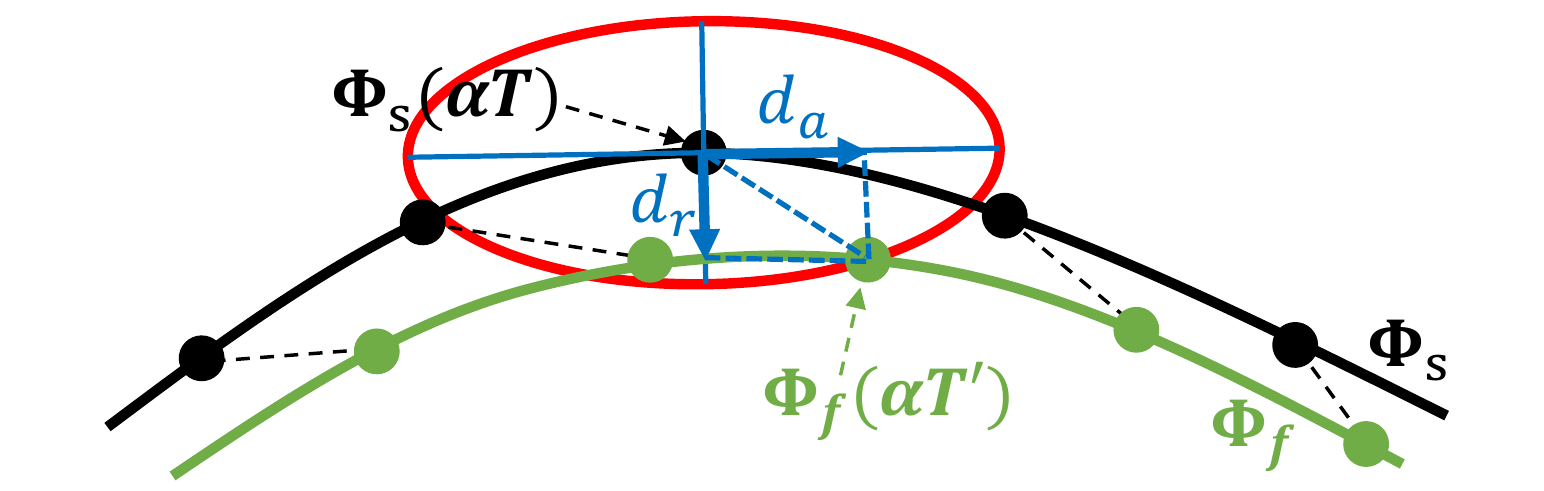}
	\captionsetup{font={small}}
	\caption{Optimizing trajectory $\mathbf{\Phi}_f$ to fit trajectory $\mathbf{\Phi}_s$ while adjusting smoothness and feasibility. Black and green dots are sample points on the trajectory. The displacement between $\mathbf{\Phi}_f(\alpha T')$ and $\mathbf{\Phi}_s(\alpha T)$ breaks down into $d_a$ and $d_r$ along two ellipse principal axes. Points at the red ellipse surface produce identical penalties. } 
	\label{pic:fitting}
\end{figure}

Firstly, as Fast-Planner \cite{boyu2019ral} does, we compute the limits exceeding ratio,
\begin{equation}
r_e = max\{ |\mathbf{V}_{i,r}/v_{m}|, \sqrt{|\mathbf{A}_{j,r}/a_{m}|}, \sqrt[3]{|\mathbf{J}_{k,r}/j_{m}|}, 1 \},
\end{equation} 
where $i\in\{ 1,\cdots, N_c-1 \}$, $j\in\{ 1,\cdots, N_c-2 \}$, $k\in\{ 1,\cdots, N_c-3 \}$ and $r\in\{x,y,z\}$ axis.
A notion with subscript $m$ represents the limitation of a derivative.
$r_e$ indicates how much we should lengthen the time allocation for $\mathbf{\Phi}_{f}$ relative to $\mathbf{\Phi}_{s}$.
Note that $\mathbf{V}_i$, $\mathbf{A}_j$ and $\mathbf{J}_k$ are inversely proportional to $\triangle t$, the square of $\triangle t$ and the cubic of $\triangle t$, respectively, from Equ.\ref{equ:v_a_j}.
Then we obtain the new time span of $\mathbf{\Phi}_{f}$
\begin{equation}
\triangle t' = r_e \triangle t.
\end{equation}

$\mathbf{\Phi}_{f}$ of time span $\triangle t'$ is initially generated under boundary constraints while maintaining the identical shape and control points number to $\mathbf{\Phi}_s$, by solving a closed-form min-least square problem.
The smoothness and feasibility are then refined by optimization.
The penalty function $J'$ formulated by linear combinations of smoothness (Sec.\ref{sec::smoothness_penalty}), feasibility (Sec.\ref{sec::feasibility_penalty}) and curve fitting (introduced later) is
\begin{equation}
\mathop{min}_{\mathbf{Q}} J' = \lambda_s J_s + \lambda_d J_d + \lambda_f J_f,
\end{equation}
where $\lambda_f$ is the weight of fitness term.

The fitting penalty function $J_f$ is formulated as the integral of anisotropic displacements from points $\mathbf{\Phi}_f (\alpha T')$ to the corresponding $\mathbf{\Phi}_s (\alpha T)$, where $T$ and $T'$ are the trajectory duration of $\mathbf{\Phi}_s$ and $\mathbf{\Phi}_f$, $\alpha \in [0,1]$.
Since the fitted curve $\mathbf{\Phi}_{s}$ is already collision-free, we assign the axial displacement of two curves with low penalty weight to relax smoothness adjustment restriction, and radial displacement with high penalty weight to avoid collision.
To achieve this, we use the spheroidal metric, shown in Fig.\ref{pic:fitting}, such that displacements at the same spheroid surface produce identical penalties.
The spheroid we use for $\mathbf{\Phi}_f(\alpha T')$ is obtained by rotating an ellipse centering at $\mathbf{\Phi}_s(\alpha T)$ about one of its principal axes, the tangent line $\dot{\mathbf{\Phi}}_s(\alpha T)$.
So the axial displacement $d_a$ and radial displacement $d_r$ can be calculated by
\begin{equation}
\begin{aligned}
d_a & = \left(\mathbf{\Phi}_{f}-\mathbf{\Phi}_{s}\right) \cdot \frac{\dot{\mathbf{\Phi}}_{s}}{\|\dot{\mathbf{\Phi}}_{s}\|}, \\
d_r & = \left\|(\mathbf{\Phi}_{f}-\mathbf{\Phi}_{s}) \times \frac{\dot{\mathbf{\Phi}}_{s}}{\|\dot{\mathbf{\Phi}}_{s}\|} \right\|.
\end{aligned}
\end{equation}

The fitness penalty function is 
\begin{equation}
\label{equ:fitness}
J_f = \int_0^{1} \left[\frac{d_a(\alpha T')^2}{a^2} + \frac{d_r(\alpha T')^2}{b^2} \right]\mathrm{d}\alpha,
\end{equation}
where $a$ and $b$ are semi-major and semi-minor axis of the ellipse, respectively.
The problem is solved by L-BFGS.

\section{Experiment Results}
\label{sec:results}

\subsection{Implementation Details}

The planning framework is summarized in Alg.\ref{alg:Rebound_Planning}.
We set the B-spline order as $p_b=3$.
The number of control points $N_c$ alters around 25, which is determined by the planning horizon (about 7m) and the initial distance interval (about 0.3m) of adjacent points.
These are empirical parameters that balance the complexity of the problem with degrees of freedom.
The time complexity is $O(N_c)$, since one control point only affects nearby segments according to the local support property of B-spline.
The complexity of L-BFGS is also linear on the same relative tolerance.
For collision-free path searching, we adopt A*, which has a good advantage that the path $\mathbf{\Gamma}$ always tends to be close to the obstacle surface naturally. 
Therefore, we can directly select $\mathbf{p}$ at $\mathbf{\Gamma}$ without obstacle surface searching.
For vector $\mathbf{R}_i$ defined in Fig.\ref{pic:3d_p_v_pairs}, it can be deduced by the property of uniform B-spline parameterization, that the $\mathbf{R}_i$ satisfies 
\begin{equation}
\mathbf{R}_i = \frac{\mathbf{Q}_{i+1} - \mathbf{Q}_{i-1}}{2\triangle t},
\end{equation}
which can be efficiently computed.
Equ.\ref{equ:fitness} is discretized to a finite number of points $\mathbf{\Phi}_f(k\triangle t')$ and $\mathbf{\Phi}_s(k\triangle t)$, where $k \in \mathbb{N}, 0 \leq k \leq \lfloor{T/\triangle t}\rfloor$.
To further enforce safety, a collision check of a circular pipe with a fixed radius around the final trajectory is performed to provide enough obstacle clearance.
The optimizer stops when no collision is detected. 
Real-world experiments are presented on the same flight platform of \cite{gao2020teach} with depth acquired by Intel RealSense D435\footnote{https://www.intelrealsense.com/depth-camera-d435/}.
Furthermore, we modify the ROS driver of Intel RealSense to enable the laser emitter strobe every other frame. 
This allows the device to output high quality depth images with the help of the emitter, and along with binocular images free from laser interference. 
The modified driver is open-sourced as well.

\setlength{\textfloatsep}{0pt}
\begin{algorithm}[t]
	\caption{Rebound Planning}
	\begin{algorithmic}[1]
		\State \textbf{Notation}: Goal $\mathcal{G}$, Environment $\mathcal{E}$, Control Point Struct $\mathbf{Q}$, Penalty $J$, Gradient $\mathbf{G}$
		\State Initialize: $\mathbf{Q} \leftarrow $ FindInit($\mathbf{Q}_{last}$, $\mathcal{G}$)
		\While{$\neg$ IsCollisionFree($\mathcal{E},\mathbf{Q}$)}
		\State CheckAndAddObstacleInfo($\mathcal{E},\mathbf{Q}$)
		\State ($J,\mathbf{G}$) $\leftarrow$ EvaluatePenalty($\mathbf{Q}$)
		\State $\mathbf{Q} \leftarrow$ OneStepOptimize($J,\mathbf{G}$)
		\EndWhile
		\If{$\neg$ IsFeasible($\mathbf{Q}$)}
		\State $\mathbf{Q} \leftarrow$ ReAllocateTime($\mathbf{Q}$) 
		\State $\mathbf{Q} \leftarrow$ CurveFittingOptimize($\mathbf{Q}$)
		\EndIf
		\State \Return $\mathbf{Q}$
	\end{algorithmic}
	\label{alg:Rebound_Planning}
\end{algorithm}

\subsection{Optimization Algorithms Comparison}
\label{sec:solver_comp}

In this section, three different optimization algorithms, including Barzilai-Borwein (BB) method, limited-memory BFGS (L-BFGS) and truncated Newton (T-NEWTON) method \cite{steiha1983truncatednewton}, are discussed. 
Specifically, each algorithm runs for 100 times independently in random maps.
All relevant parameters including boundary constraints, time allocation, decision variables initialization, and random seeds, are set identical for different algorithms. 
The data about success rate, computation time and numbers of objective function evaluations are recorded.
Only the successful cases are counted due to the data in failed cases is meaningless.
The associated results are shown in Tab.\ref{tab:solver_comparison}, which states that L-BFGS significantly outperforms the other two algorithms.
L-BFGS characterizes a type of approximation by means of second order Taylor expansions, which is suitable for optimizing the objective function described in Sec.\ref{sec::problem_optimization}.
Truncated Newton method approximates the second order optimization direction $\mathbf{H}^{-1} \nabla \mathbf{f}_k$ as well. However, too many objective function evaluations increase the optimization time.
BB-method estimates the Hessian as a scalar $\lambda$ times $\mathbf{I}$.
Nevertheless, the insufficient estimation of Hessian still leads to a low convergence rate.

\renewcommand{\arraystretch}{1.1} 
\begin{table}[ptb]   
	\centering  
	\fontsize{6.5}{8}\selectfont  
	\begin{threeparttable}  
		\caption{Optimization Algorithms Comparison}  
		\label{tab:solver_comparison}  
		\begin{tabular}{cccccccc}  
			\toprule  
			\multirow{2.6}{*}{Algorithm}&\multirow{2.6}{*}{Success Rate}&
			\multicolumn{3}{c}{Time(ms)}&\multicolumn{3}{c}{Function Evaluations}\cr  
			\cmidrule(lr){3-5} \cmidrule(lr){6-8}  
			& &Min&Avg&Max&Min&Avg&Max\cr  
			\midrule  
			BB & 0.86 & 0.21 & 0.50 & 1.14 & 108 & 268.2 & 508 \cr
			T-NEWTON & 0.62 & 0.3 & 0.79 & 3.59 & 109 & 344.29 & 702 \cr
			L-BFGS & \bf 0.89 & \bf 0.17 & \bf 0.37 & \bf 0.80 & \bf 22 & \bf 79.04 & \bf 182 \cr
			\bottomrule  
		\end{tabular}  
	\end{threeparttable} 
\end{table}

\renewcommand{\arraystretch}{1.1} 
\begin{table}[pt]  
	\centering  
	\fontsize{6.5}{8}\selectfont  
	\begin{threeparttable}  
		\caption{ESDF/ESDF-free Methods Comparison}  
		\label{tab:ESDF_comparison}  
		\begin{tabular}{cp{0.8cm}<{\centering}p{0.6cm}<{\centering}p{0.5cm}<{\centering}p{0.5cm}<{\centering}p{0.8cm}<{\centering}p{0.6cm}<{\centering}p{0.6cm}<{\centering}}  
			\toprule  
			\multirow{2.6}{*}{Method}&
			\multirow{2.6}{*}{\shortstack{Success\\Rate}}&
			\multirow{2.6}{*}{Energy}&
			\multicolumn{2}{c}{Velocity(m/s)}&
			\multicolumn{3}{c}{Time(ms)}\cr  
			\cmidrule(lr){4-5} \cmidrule(lr){6-8}  
			& & &Avg&Max&Optimize&ESDF&Total\cr  
			\midrule  
			EGO &\bf 0.89&49.92&2.12&\bf 2.24&\bf 0.37&\bf /&\bf 0.37 \cr
			ENI & 0.69&\textcolor[rgb]{0.5,0.5,0.5}{35.55}&\textcolor[rgb]{0.5,0.5,0.5}{2.09}&\textcolor[rgb]{0.5,0.5,0.5}{2.23}&\textcolor[rgb]{0.5,0.5,0.5}{0.43}&\textcolor[rgb]{0.5,0.5,0.5}{5.03}&\textcolor[rgb]{0.5,0.5,0.5}{5.46} \cr
			EI &\bf 0.89&\bf 42.27&\bf 2.11&2.36&0.48&5.07&5.55 \cr
			\bottomrule  
		\end{tabular}  
	\end{threeparttable}  
	\vspace{-1.5cm}
\end{table}
 
\subsection{Trajectory Generation With \& Without ESDF}
\label{sec:ESDF_comp}

We use the same setting as Sec.\ref{sec:solver_comp} to perform this comparison.
On account of the low success rate explained in \cite{boyu2019ral} when using straight line initialization for an ESDF-based trajectory generator, we adopt a collision-free initialization.
Comparison results are in Tab.\ref{tab:ESDF_comparison}.

\begin{figure*}[t]
\centering
\includegraphics[width=1\linewidth]{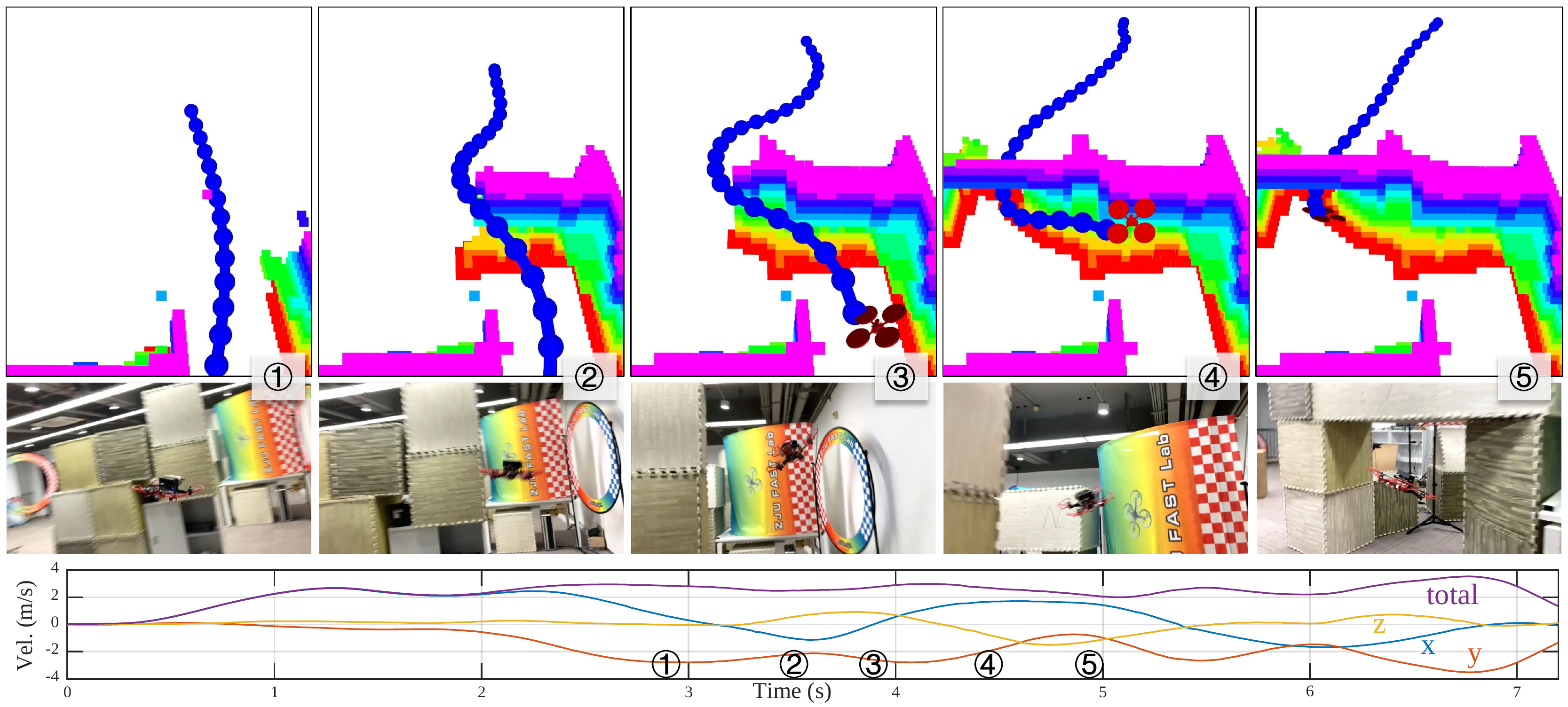}
\captionsetup{font={small}}
\caption{ Visualization of local trajectory planning over a short period of time with velocity profile. }
\label{pic:traj_gen}
\vspace{-0.8cm}
\end{figure*}

For clarity, ESDF-based methods with and without collision-free initialization are abbreviated as \textit{EI} and \textit{ENI}.
This comparison gives that the proposed EGO algorithm achieves a comparable success rate to ESDF-based methods with collision-free initialization.
However, trajectory energy (jerk integral) produced by EGO is slightly higher.
This happens because the control points of EGO which contain more than one $\{\mathbf{p},\mathbf{v}\}$ pair produce stronger trajectory deformation force than EI does, as described in Sec.\ref{sec::colli_fun}.
On the other hand, stronger force accelerates the convergence procedure, resulting in shorter optimization time.
Some statistics of ENI (shown in gray) can be less convincing because ENI tests can only succeed in fewer challenge cases where the resulting trajectories are naturally smoother with less energy cost and lower velocity, compared to EI and EGO.
Something noteworthy is that although the ESDF updating size is reduced to $10 \times 4 \times 2~m^3$ with $0.1m$ resolution for a $9m$ trajectory, the ESDF updating still takes up a majority of the computation time.

\subsection{Multiple Planners Comparison}
\label{sec:plan_comp}

\begin{figure}[t]
	\vspace{0.5cm}
	\centering
	\includegraphics[width=1\linewidth]{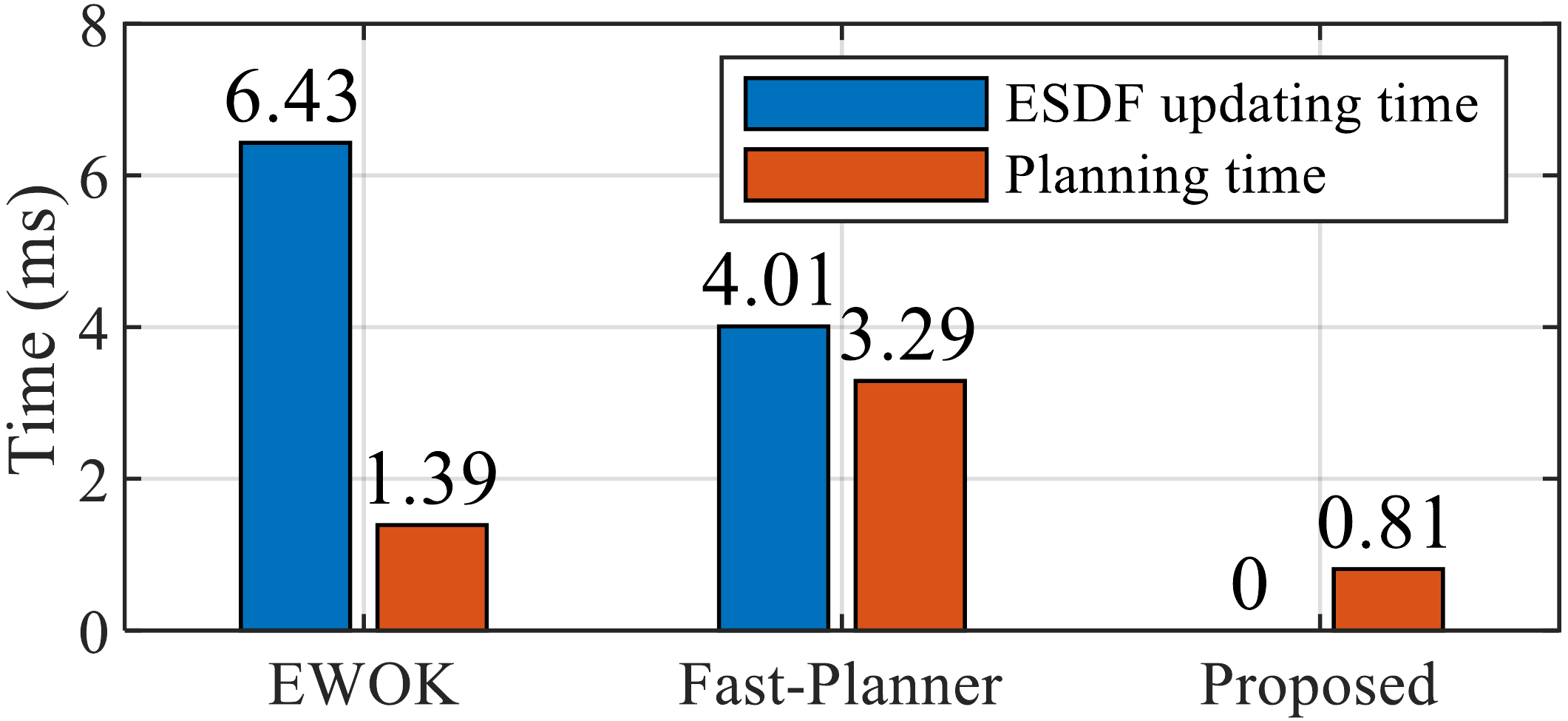}
	\captionsetup{font={small}}
	\caption{ Comparison of the proposed EGO-Planner against two SOTA planners with default parameters. }
	\label{pic:plan_cmp}
	\vspace{-0.6cm}
\end{figure}

\begin{figure}[t]
	\vspace{0.6cm}
	\centering
	\includegraphics[width=1\linewidth]{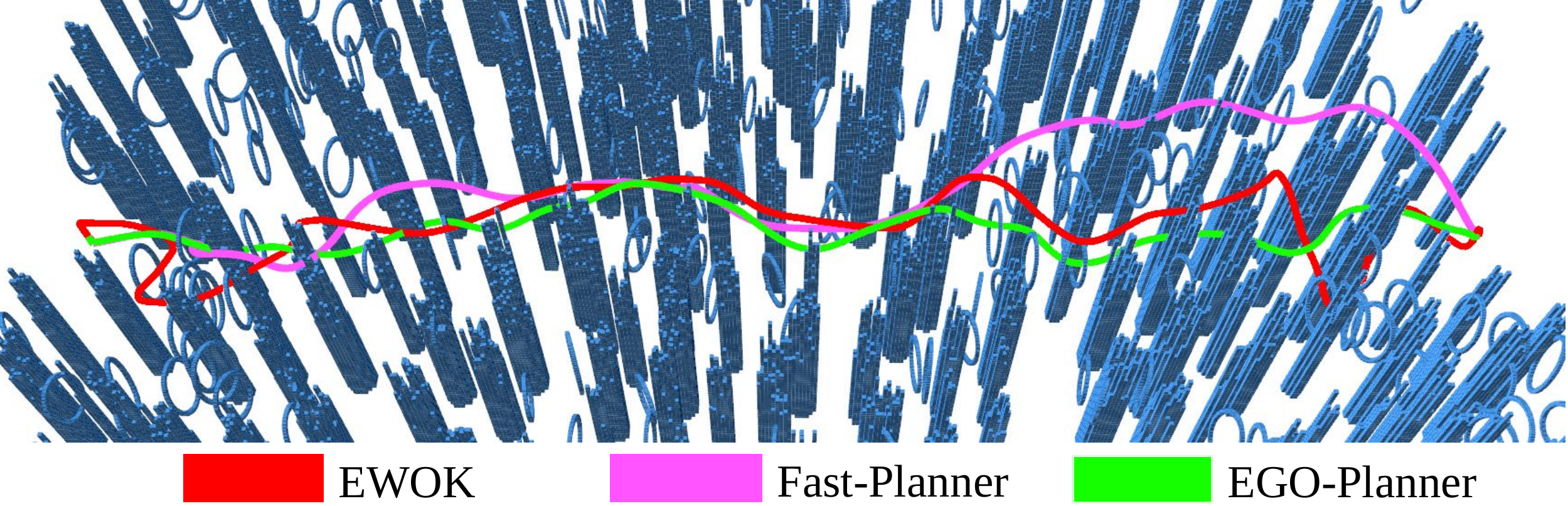}
	\captionsetup{font={small}}
	\caption{ Trajectory visualization in simulation. }
	\label{pic:traj_cmp}
	\vspace{-0.6cm}
\end{figure}

We compare the proposed planner with two state-of-the-art methods, Fast-Planner\cite{boyu2019ral} and EWOK\cite{Usenko2017ewok}, which utilize ESDF to evaluate obstacle distance and gradient.
Each planner runs for ten times of different obstacle densities from the same starts to ends.
The average performance statistics and the ESDF computation time are shown in Tab.\ref{tab:plan_cmp2} and Fig.\ref{pic:plan_cmp}.
Trajectories generated by three methods on a map of 0.5 obstacles/$m^2$ are illustrated in Fig.\ref{pic:traj_cmp}.

From Tab.\ref{tab:plan_cmp2} we conclude that the proposed method achieves shorter flight time and trajectory length but ends up in higher energy cost compared to Fast-Planner.
This is mainly caused by the front-end kinodynamic path searching in \cite{boyu2019ral}.
EWOK suffers twisty trajectories in dense environments, since the objective function contains exponential terms, which leads to unstable convergence in optimization.
Furthermore, we conclude that a lot of computation time without ESDF updating is saved by the proposed method.

\begin{table}[ptb]
	\vspace{0.5cm}
	\centering
	\caption{Planners Comparison}  
	\begin{tabular}{c|ccccc}
		\toprule
		Planner & $t$(s) & Length & Energy & $t_\text{ESDF}$(ms) & $t_\text{plan}$(ms) \\
		\midrule
		EWOK & 31.00 & 59.05 & 246.12 & 6.43 & 1.39        \\
		Fast-Planner & 30.76 & 45.18 & \bf 135.21 & 4.01 & 3.29        \\
		EGO-Planner & \bf 24.38 & \bf 42.24 &196.64 & \bf / & \bf 0.81       \\
		\bottomrule
	\end{tabular}
	\label{tab:plan_cmp2}
	\vspace{-1.0cm}
\end{table} 

\begin{figure*}[t]
	\centering
	\includegraphics[width=1\linewidth]{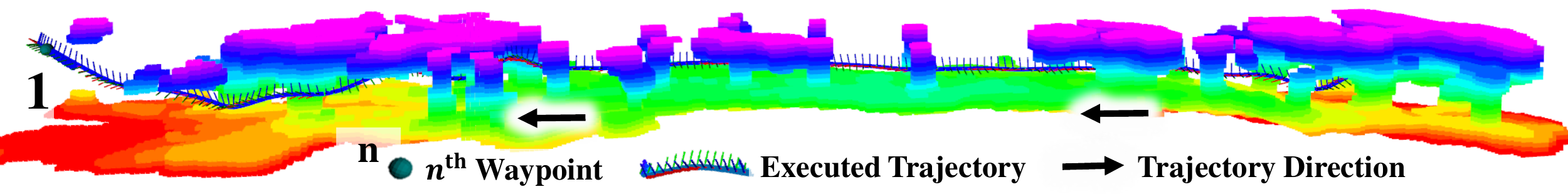}
	\captionsetup{font={small}}
	\caption{ Trajectory of an outdoor experiment in a forest. }
	\label{pic:traj_outdoor}
\end{figure*}

\subsection{Real-world Experiments}
\label{sec:realworld_exp}
\begin{figure}[t]
	\centering
	\includegraphics[width=1\linewidth]{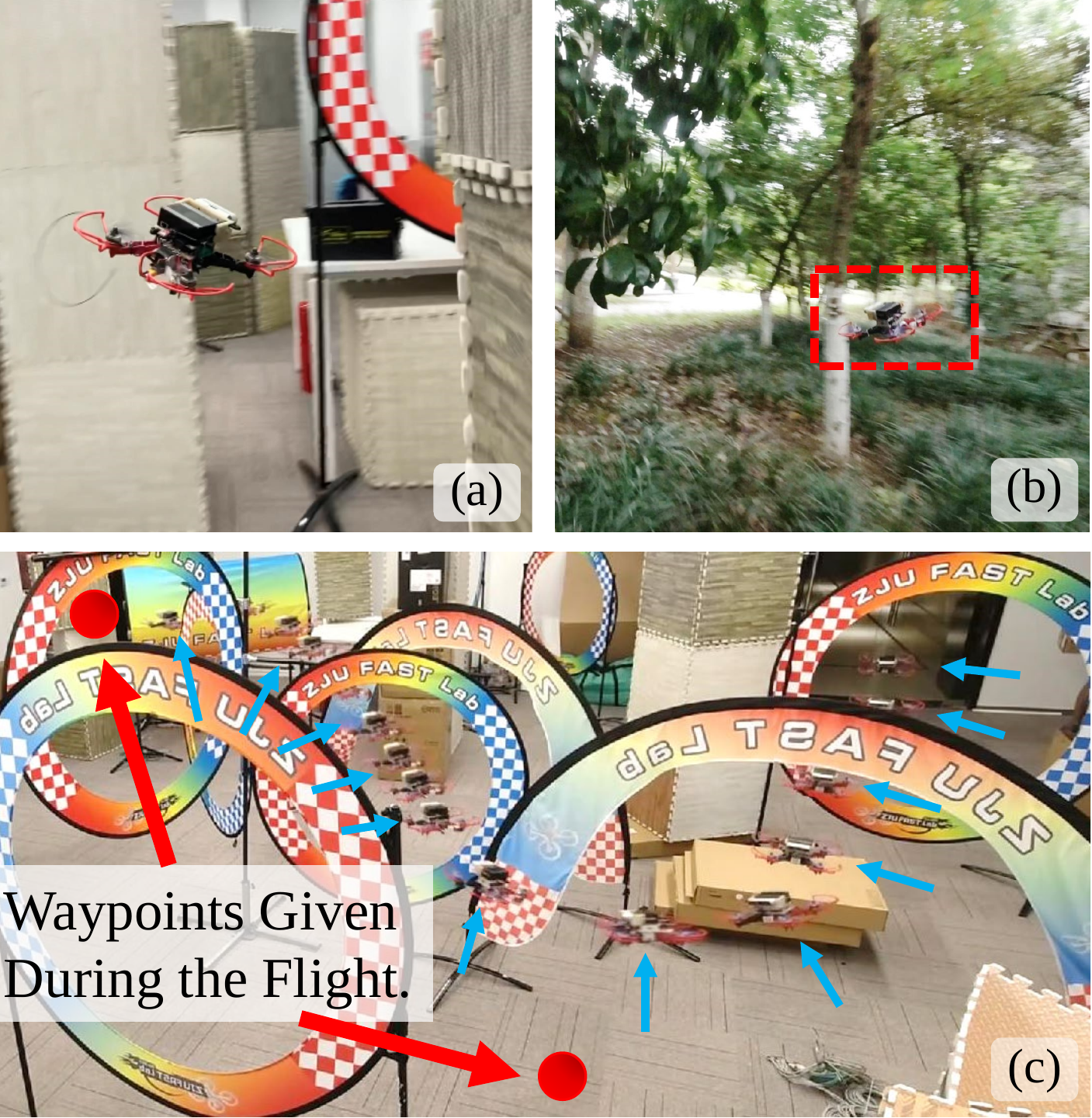}
	\captionsetup{font={small}}
	\caption{ Real-world experiments. a) An indoor test. b) An outdoor test. c) Composite snapshots of indoor flights. }
	\label{pic:realworld_exp}
\end{figure}

\begin{figure}[t]
\centering
\includegraphics[width=1\linewidth]{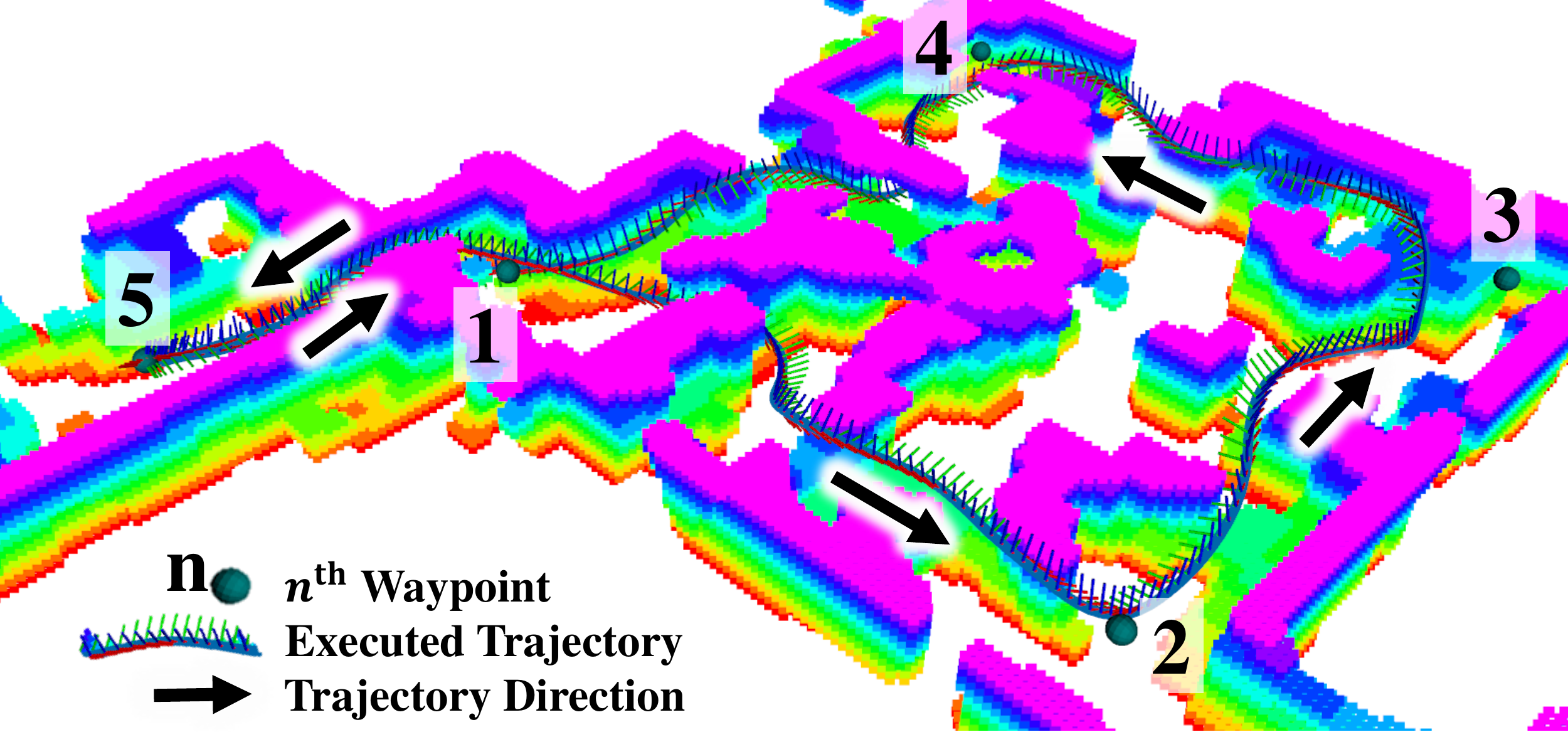}
\captionsetup{font={small}}
\caption{ Trajectory of an indoor experiment }
\label{pic:traj_indoor}
\end{figure}

We present several experiments in cluttered unknown environments with limited camera FOV.
One experiment is to fly by waypoints given in advance. 
In this experiment, the drone starts from a small office room, passes through the door, flies around in a big cluttered room, and then returns to the office, as illustrated in Fig.\ref{pic:realworld_exp}a and Fig.\ref{pic:traj_indoor}.
The narrowest passage of indoor experiments is less than one meter as shown in Fig.\ref{pic:traj_gen}. By contrast, the drone reaches $3.56m/s$ in such a cluttered environment.

Another indoor experiment is to chase goals arbitrarily and abruptly given during the flight, as shown in Fig.\ref{pic:realworld_exp}c.
In this test, limited FOV puts greater challenges that a feasible trajectory must be generated immediately once a new goal is received or collision threat is detected.
Thus, this experiment validates that the proposed planner is capable of performing aggressive flight on the premise of feasibility.

In the outdoor experiments, the drone flies through a forest of massive trees and low bushes, as shown in Fig.\ref{pic:realworld_exp}b and Fig.\ref{pic:traj_outdoor}.
Although the wild airflow around the drone causes swinging of the branches and leaves, making the map less reliable, the drone still reaches a speed above $3m/s$.
Therefore, the proposed planner can tackle both experimental and field environments.
We refer readers to the video\footnote{https://youtu.be/UKoaGW7t7Dk} for more information.

\section{Conclusion and Future Work}
\label{sec:conclusion_F.W.}

In this paper, we investigate the necessity of ESDF for gradient-based trajectory planning and propose an ESDF-free local planner.
It achieves comparable performance to some state-of-the-art ESDF-based planners but reduces computation time for over an order of magnitude.
Benchmark comparisons and real-world experiments validate that it is robust and highly efficient.

The proposed method still has some flaws, which are the local minimum introduced by A* search and the conservative trajectories introduced by unified time re-allocation.
Therefore, we will work on performing topological planning to escape the local minimum and re-formulating the problem to generate near-optimal trajectories.
The planner is designed for static environments and can tackle slowly moving obstacles (below 0.5m/s) without any modification.
We will work on dynamic environment navigation by moving object detection and topological planning in the future.

\vspace{1.0cm}

\newlength{\bibitemsep}\setlength{\bibitemsep}{0.0\baselineskip}
\newlength{\bibparskip}\setlength{\bibparskip}{0pt}
\let\oldthebibliography\thebibliography
\renewcommand\thebibliography[1]{%
	\oldthebibliography{#1}%
	\setlength{\parskip}{\bibitemsep}%
	\setlength{\itemsep}{\bibparskip}%
}
\bibliography{xin_ral2020}
\end{document}